\documentclass[conference]{IEEEtran}
\IEEEoverridecommandlockouts
\usepackage{cite}
\usepackage{booktabs}
\usepackage{multirow}
\usepackage{amsmath,amssymb,amsfonts}
\usepackage{algorithmic}
\usepackage{graphicx}
\usepackage{url}
\usepackage{textcomp}
\usepackage{eso-pic}
\usepackage{xcolor}
\usepackage{tikz}
\def\BibTeX{{\rm B\kern-.05em{\sc i\kern-.025em b}\kern-.08em
    T\kern-.1667em\lower.7ex\hbox{E}\kern-.125emX}}

\newcommand\copyrighttext{%
  \footnotesize © 2023 IEEE. Personal use of this material is permitted.
  Permission from IEEE must be obtained for all other uses, in any current or future
  media, including reprinting/republishing this material for advertising or promotional
  purposes, creating new collective works, for resale or redistribution to servers or
  lists, or reuse of any copyrighted component of this work in other works.
  DOI: 10.1109/ICT4S58814.2023.00015. Available at:\\ \url{https://ieeexplore.ieee.org/document/10292174}}
\newcommand\copyrightnotice{%
\begin{tikzpicture}[remember picture,overlay]
\node[anchor=south,yshift=10pt] at (current page.south) {\fbox{\parbox{\dimexpr\textwidth-\fboxsep-\fboxrule\relax}{\copyrighttext}}};
\end{tikzpicture}%
}
    
\begin{document}

\title{Energy cost and machine learning accuracy impact of k-anonymisation and synthetic data techniques}

\author{\IEEEauthorblockN{Pepijn de Reus}
\IEEEauthorblockA{\textit{Master Artificial Intelligence} \\
\textit{University of Amsterdam}\\
Amsterdam, The Netherlands \\
p.dereus@uva.nl}
\and
\IEEEauthorblockN{Ana Oprescu}
\IEEEauthorblockA{\textit{Complex Cyber Infrastructure} \\
\textit{University of Amsterdam}\\
Amsterdam, The Netherlands \\
a.m.oprescu@uva.nl}
\and
\IEEEauthorblockN{Koen van Elsen}
\IEEEauthorblockA{\textit{Institute for Informatics} \\
\textit{University of Amsterdam}\\
Amsterdam, The Netherlands \\
k.m.j.vanelsen@uva.nl}
}

\maketitle

\begin{abstract}
To address increasing societal concerns regarding privacy and climate, the EU adopted the General Data Protection Regulation (GDPR) and committed to the Green Deal. Considerable research studied the energy efficiency of software and the accuracy of machine learning models trained on anonymised data sets. Recent work began exploring the impact of privacy-enhancing techniques (PET) on \textit{both} the energy consumption and accuracy of the machine learning models, focusing on $k$-anonymity. As synthetic data is becoming an increasingly popular PET, this paper analyses the energy consumption and accuracy of two phases: a) applying privacy-enhancing techniques to the concerned data set, b) training the models on the concerned privacy-enhanced data set. We use two privacy-enhancing techniques: k-anonymisation (using generalisation and suppression) and synthetic data, and three machine-learning models. Each model is trained on each privacy-enhanced data set. Our results show that models trained on $k$-anonymised data consume less energy than models trained on the original data, with a similar performance regarding accuracy. Models trained on synthetic data have a similar energy consumption and a similar to lower accuracy compared to models trained on the original data.\\
\end{abstract}


\copyrightnotice

\section{Introduction}
To address climate change the European Commission has set a goal of reducing net carbon emission to zero in 2050~\cite{GreenDeal2020}. The amount of publications on Artificial Intelligence (AI) has increased more than fivefold over the last decade~\cite{Npublications}. Early studies already warned of the dangers of using digitalisation without rebound considerations regarding the environment~\cite{coroamua2019digital} and identified the need for a more fine-grained analysis of digital processes with respect to their ecological footprint~\cite{kopp2019climate}. Nowadays, several international initiatives, such as the EU Green Deal, aim to reduce carbon emissions. With this goal in mind, the EU aims to make data centres and ICT infrastructures climate-neutral by 2030 and aims to make use of artificial intelligence and other digital technologies to reduce the impact of climate change~\cite{EU_2020}. This has encouraged more research with a focus on energy consumption within machine learning~\cite{garcia2017energy, garcia2019estimation, li2016evaluating, verdecchia2022data}.

Apart from the energy aspect there is growing concern about privacy amongst citizens of all ages~\cite{elueze2018privacy}. To ensure the privacy of those whose data are collected the General Data Protection Regulation (GDPR) has been adopted in Europe in 2016. The GDPR regulates that all citizens in Europe have control over their personal data~\cite{EUdataregulations2018}. There is one exception to the GDPR; it does not apply to anonymised data. In the GDPR, anonymous data is defined as: "information which does not relate to an identified or identifiable natural person or to personal data rendered anonymous in such a manner that the data subject is not or no longer identifiable."~\cite{EUAnony2018}. Therefore it is interesting to look into methods to anonymise data such that data can be shared without GDPR constraints, especially in light of the upcoming Data Act~\footnote{\url{https://ec.europa.eu/commission/presscorner/detail/en/ip_22_1113}}. 

One common method for enhancing privacy of data is $k$-anonymity, using either generalisation and suppression~\cite{sweeney2002k} or micro-aggregation~\cite{defays1998masking}. Apart from $k$-anonymity, synthetic data is becoming increasingly popular as a privacy-enhancing technique~\cite{machanavajjhala2008privacy}. Synthetic data is unique as it mimics the original data in correlations and properties, yet it does not contain the original data~\cite{rubin1993statistical}.

Combining these societal concerns and trends raises the question of what privacy-enhancing algorithms add to the energy consumption of regular machine learning tasks. And how do these algorithms relate to the accuracy of machine learning models? In this paper, we analyse both aspects for both $k$-anonymity and synthetic data. Previous research evaluated the impact of the $k$-value~\cite{Kouwenberg_2020,Madan_2021,Bhati_2021}, and of synthetic data~\cite{ping2017datasynthesizer, hittmeir2019utility} on the accuracy of the resulting machine learning models. While there is work analysing both the energy cost and accuracy impact of the k-value~\cite{oprescu2022energyk}, to the best of our knowledge, no work has been conducted on comparing synthetic data with $k$-anonymised data looking at both accuracy and energy consumption. Understanding this impact could help users make an informed decision between energy consumption, accuracy, and anonymity.

\subsection{Research questions}
In the context of considering both an increased use of data and its climate impact, one exploratory goal of this research is to understand which method would be preferred when using data protection: $k$-anonymity or synthetic data?

In this paper we examine two directly related questions:
\begin{enumerate}
    \item[\textbf{RQ1:}] Which method yields the least loss of accuracy?
    \item[\textbf{RQ2:}] Which method is the least energy consuming?
\end{enumerate}

\subsection*{Outline}
We briefly review background knowledge in Section~\ref{sec:background} and we explain our research method in Section~\ref{sec:method}. In Section~\ref{sec:experiments} we describe the experimental set-up. The results are shown in Section~\ref{sec:results} and discussed in Section~\ref{sec:discussion}. In Section~\ref{sec:rel-work} we relate similar efforts to our research and we conclude in Section~\ref{sec:conclusion}.

\section{Background}
\label{sec:background}
We briefly present several concepts that are key for our research. 
\subsection{$k$-anonymity}
A well known property to define anonymised data is $k$-anonymity, which was first introduced by Sweeney~\cite{sweeney2002k}. In this paper Sweeney proved that anonymisation by removal of identifying attributes such as names was insufficient to ensure privacy. By linking two publicly available, anonymised, data sets Sweeney was able to identify 87\% of the data subjects using a combination of ZIP code and date of birth~\cite{sweeney2002k}. Following this Sweeney introduced $k$-anonymity to protect data subject's privacy. Shortly put, a data set with a $k$-anonymity of 5 ensures that for each data subject in the set, at least 5 other data subjects share the exact same properties. These 6 data subjects would thus be indistinguishable from each other. To come to this $k$-anonymity, all attributes should be classified as one of four data types:
\begin{itemize}
    \item \textbf{Insensitive} attributes are deemed unimportant for privacy and will remain unaltered.
    \item \textbf{Sensitive} attributes are important for the subject (think of political view).
    \item \textbf{Identifying} attributes such as names are directly linked to a person and will be removed from the data.
    \item \textbf{Quasi-identifying} attributes refer to the data that could compromise privacy when linked with other attributes or data sets.
\end{itemize}

\begin{figure}[h]
    \centering
    \includegraphics[width=0.42\textwidth]{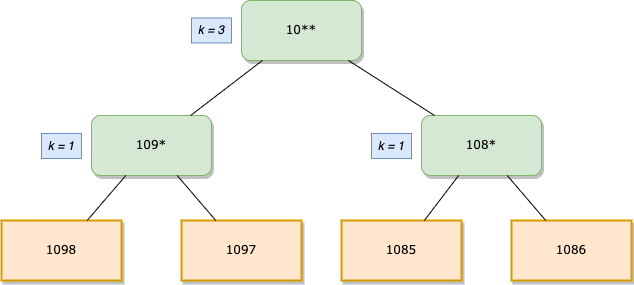}
    \caption{Generalisation and suppression for ZIP codes. In the orange boxes we see the original values, green boxes represent the anonymised boxes using generalisation and suppression. Next to the anonymised boxes we have the corresponding $k$-value.}
    \label{Generalisation}
\end{figure}

Once the attributes of a data set are labeled using the data types above and our value for $k$ is picked, we can use generalisation and suppression to alter our data such that it suffices to this $k$-anonymity~\cite{sweeney2002k}. It could be that data that is labeled as quasi-identifying is in fact a unique tracker because one data subject has a unique value for this attribute. A known example is that a university has only one dean. If a university were to publish the salaries, the function of a dean is a unique tracker. Even though functions in general are not per se identifying, e.g. there could be hundreds of teachers. For generalisation and suppression we need hierarchies to group our data, as can be seen for ZIP codes in Figure \ref{Generalisation}.
The first level of our hierarchy would be to suppress the last digit, after which we could suppress all but the first digit. The algorithm will try to leave as much data as is, but will increase the hierarchy if the attribute prevents the data set from obtaining a certain $k$-anonymity. These hierarchies are increased until the desired $k$-anonymity is reached or until the entire attribute is suppressed.

\subsection{Synthetic data}
An increasingly popular method~\cite{hernandez2022synthetic} to share data without comprising the subjects' privacy is synthetic data. Synthetic data is a form of AI that learns the properties, covariance and distributions of a data set. Using this information, a new, synthetic, data set can be constructed with very similar statistics to the original data set such that the performance of machine learning on the synthetic data set is similar to that of the original data set~\cite{abadi2016deep}. By making sure that each data subject within the synthetic data is not present in the actual data we can publish the data set without compromising the data subject's privacy~\cite{rubin1993statistical}.

\subsection{Measuring energy consumption}
Processors of Intel from 2012 and later have Running Average Power Limiters (RAPL) energy sensors in their CPUs~\cite{desrochers2016validation, hahnel2012measuring}. The RAPL sensors measure the energy consumption of the CPU and DRAM of the Intel chip. An advantage of RAPL is that the user can select which parts of the chip should be measured~\cite{khan2018rapl,Garcia_2019}, e.g. solely the CPU or the DRAM energy consumption. 

\section{Research method}
\label{sec:method}
In order to answer the research questions a setup was designed to measure both the energy consumption and the impact on the accuracy of the selected machine learning techniques. The input of this setup is the data, the values of k in the $k$-anonymisation process and lastly the hyperparameters for the synthetic data and the machine learning techniques. This will output the total energy consumption of the setup and the accuracy score for each ML technique based on the input data.

Firstly the data is preprocessed, after which either the synthetic data generator or the anonymisation algorithm is used to obtain the new data. This data is then fed to the machine learning techniques which will yield an accuracy for each technique. This will give insights to answer the research question regarding  loss of accuracy (\textbf{RQ1}). Over this entire process we measure the energy consumption by reading the RAPL counters, which helps to answer our research question for the energy consumption (\textbf{RQ2}).

\subsection{Data}
To answer our research questions we require data sets that contain personal information, preferably with unique trackers. In general, such data sets are hard to obtain because of their privacy sensitive character. A good alternative is to use data sets from the UC Irvine Machine Learning repository, which contains many public data sets for machine learning research~\cite{BacheLichman2013}. Preferably the data sets differ in size and in the type of data they contain so that we can diversify our findings. If a data set contains missing information we choose to delete these rows as is common in literature~\cite{escobar2021quality}. After deleting the rows with missing information we say our data is cleaned. 

Following this data cleaning we anonymise the data using $k$-anonymity using generalisation and suppression. Micro-aggregation is left out as recent work suggests that anonymisation by generalisation and suppression has less loss of accuracy~\cite{oprescu2022energyk}. As explained in the background we first wish to, manually, label our attributes as either insensitive, sensitive, identifying or quasi-identifying. Insensitive information will remain unaltered and identifying information will be suppressed, though we do not expect to find identifying attributes in public data sets. To anonymise the data we will use the $k$-values of 3, 10 and 27 as earlier work shows that these values could lead to improved accuracy scores when compared to the baseline \cite{misdorp2022}. The suppression limit is set to 20\% as is the default setting \cite{prasser2014arx}. The anonymised output data will then be used to train the machine learning techniques, after which the energy consumption and accuracy of the techniques will be compared to the baseline. 

The generator for the synthetic data will also use the cleaned data to analyse the data features and structures. By computing the probability functions and the correlations between the attributes it will output a synthetic data set. As for the anonymised data the machine learning techniques will then be trained on this synthetic data, where after the energy consumption and accuracy of the techniques will be evaluated to the baseline.

\subsection{Machine Learning}
To evaluate the impact of using $k$-anonymity and synthetic data on machine learning, three common techniques are used:
\begin{itemize}
    \item \textbf{k-nearest neighbours} groups data points based on shared features, assuming that data from the same class lies close to each other \cite{ray2019quick}.
    \item \textbf{Logistic Regression} is a common technique for classification problems \cite{hao2019machine}. It computes the probability of each class and then picks the class with the highest probability. 
    \item \textbf{Neural networks} are used for a wide variety of problems, e.g. regression and classification. Neural networks are computationally more expensive and handle noise better than most other ML techniques \cite{ray2019quick}. 
\end{itemize}

To evaluate these machine learning models, we compute the accuracy. The accuracy consists of the correctly classified units divided by the total amount of data units as implemented in sklearn and keras sequential. Since the setup of all models will be similar for the different data sets, the possibility for variations to come from anything other than the data are excluded. In this way we can measure the information loss of anonymised and synthetic data by analysing the accuracy of the machine learning models.

\begin{figure}[h]
    \centering
    \includegraphics[width=0.48\textwidth]{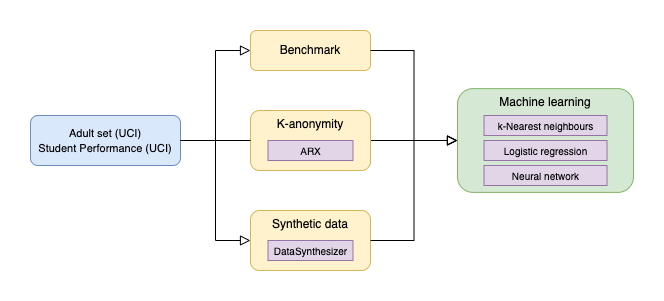}
    \caption{General overview of the experiment. In blue we see the cleaned input data, in yellow we see the anonymisation method where the purple bar specifies the used method. The green box represents the machine learning, where likewise the techniques are specified in the purple bars.}
    \label{Overview}
\end{figure}

Following the information above, we summarise the set-up in Figure \ref{Overview}.

\subsection{Measuring energy consumption}
There are two options to measure the energy consumption of our code; hardware-based or software-based \cite{khan2018rapl}. The hardware-based method, also known as power plug method, is an external device between the socket and the used machine. This device is very accurate, but can measure only the entire energy input to the device. The software-based solution uses RAPL. RAPL is a built-in tool for Intel processors and measures the energy consumption for both the CPU and DRAM. Measurements that compare RAPL with the power plug method show a correlation of 0.99 \cite{khan2018rapl}, so the influence of RAPL on the total energy consumption may be neglected. To reduce the chance of outliers we perform 10 measurements and take the weighted average. Additionally, we perform 10 idle measurements as well. In these measurements we compute the energy consumption of one second sleep in the code. This idle measurement will be averaged again and deducted from the other measurements to obtain a measurement solely based upon the used code.

\section{Experimental set-up}
\label{sec:experiments}
As RAPL is not available for macOS, even when using virtual machines, we run the code on a server instead. We run the experiments on an Ubuntu 22.04 LTS, with an 8x Intel CPU E5-1620 3.5GHz and 16 GB (4 *sk 4GB) DDR4 2400 MT/s. We used ARX 3.9.0, DataSynthesizer v0.1.11, pyRAPL 0.2.3.1, Python 3.10.4 and Java 11.0.15. The machine learning models were implemented using the sklearn (Logistic Regression and k-nearest neighbours) and Tensorflow (neural network) Python libraries. Our code is publicly available on GitHub~\footnote{\url{https://github.com/PepijndeReus/Privacy-Enhancing-ML}}.

We use two known data sets from the UC Irvine Machine Learning Repository~\cite{BacheLichman2013}, the Adult data set because it is seen as a benchmark set for k-nearest neighbours~\cite{friedman2008providing} and the Student Performance set. The Student Performance set has a different structure, as can be seen in Table~\ref{PaperDataOverview}. We use a set with a different structure as this could help gaining insight in how far these findings extend to different kinds of data sets. However to keep as many parameters equal we modify the Student Performance set such that it will also become a binary prediction task, as seen in earlier literature~\cite{cortez2008using}. Furthermore, we use the Portuguese class as this contains more instances ($n=649$) compared to the mathematics class ($n=395$).
\begin{table}[h]
\centering
\caption{Overview of the structure of the selected data sets.}
\label{PaperDataOverview}
\begin{tabular}{lrrc}\toprule
\multicolumn{2}{r}{\# entries} & \# attributes & Prediction \\
\cmidrule(r){2-4}
Adult               & 48 842     & 15            &  Binary     \\ 
Student Performance & 649       & 33            &  Value (0-20)      \\ \bottomrule
\end{tabular}
\end{table}

\subsection{k-anonymity}
To create the data sets that have $k$-anonymity we run our data trough the ARX library~\cite{prasser2014arx}, as this is often used in research to generate anonymised data and is open-source~\cite{oprescu2022energyk}. ARX requires predefined hierarchies to use generalisation and suppression. We use a similar hierarchy as in earlier work~\cite{oprescu2022energyk}, the hyperparameters are available on GitHub.

\subsection{Synthetic data}
For the creation of synthetic data we use the DataSynthesizer by Ping et al.~\cite{ping2017datasynthesizer}, as literature suggests that this method has the lowest information loss compared to other modules~\cite{hittmeir2019utility}. To minimise randomness the parameter for differential privacy is switched off and the degree of the Bayesian network is set to 2 as in the DataSynthesizer examples~\cite{ping2017datasynthesizer}. The number of instances that will be created is set to the length of the data set as shown in Table~\ref{PaperDataOverview}. As with $k$-anonymity we provide the hyperparameters on GitHub.

\section{Results}
\label{sec:results}
We present the results in the following order: A) The privacy-enhancing processes; B) Energy consumption of the models; C) Accuracy of the models.

\subsection{Privacy-enhancing processes}
In Table~\ref{PaperAnonymisationEnergy} we present the run time and energy cost to obtain the anonymised data. The table shows the deviation for each method compared to the benchmark in percentages. The deviation is computed as an average over 10 individual samples for each method. Table~\ref{PaperSuppressedData} shows the percentage of data that is suppressed after obtaining anonymised data via generalisation and suppression.

\begin{table}[h]
\centering

\caption{The run time and energy consumption for the algorithms used to obtain the anonymised data. The $k$-values specify the value given as input to the ARX algorithm, the synthetic data reports the measurements by the DataSynthesizer.}

\begin{tabular}{lcccl}
\toprule
& \multicolumn{2}{c}{\textbf{Adult data set}} & \multicolumn{2}{c}{\textbf{Student performance set}} \\
 \cmidrule(r){2-3} \cmidrule(r){4-5}
$k$=3 & 32.89s & 1013.52J & 33.49s & 1069.26J \\
$k$=10 & 31.27s & 1011.56J & 33.84s & 1069.09J \\
$k$=27 & 31.31s & 1012.60J & 34.06s & 1074.30J \\
Synthetic data & 67.48s & 3861.79J & 72.37s & 4227.54J \\
\bottomrule
\end{tabular}
\label{PaperAnonymisationEnergy}
\end{table}

\begin{table}[h]
\centering
\caption{The percentage of suppressed data for the Adult set and Student Performance with respect to the obtained $k$-anonymity. The percentage is defined as suppressed cells divided by the total amount of cells.}
\label{PaperSuppressedData}
\begin{tabular}{lcc}
\toprule
\multicolumn{3}{c}{\textbf{Suppressed data}} \\
\cmidrule(r){2-3}
& Adult set & Student Performance set \\
$k$=3 & 19\% & 74\% \\
$k$=10 & 28\% & 80\% \\
$k$=27 & 32\% & 84\% \\
\bottomrule
\end{tabular}

\end{table}

\subsection{Energy consumption of training the models}
\begin{figure*}[h]
    \centering
    \includegraphics[width=0.85\textwidth]{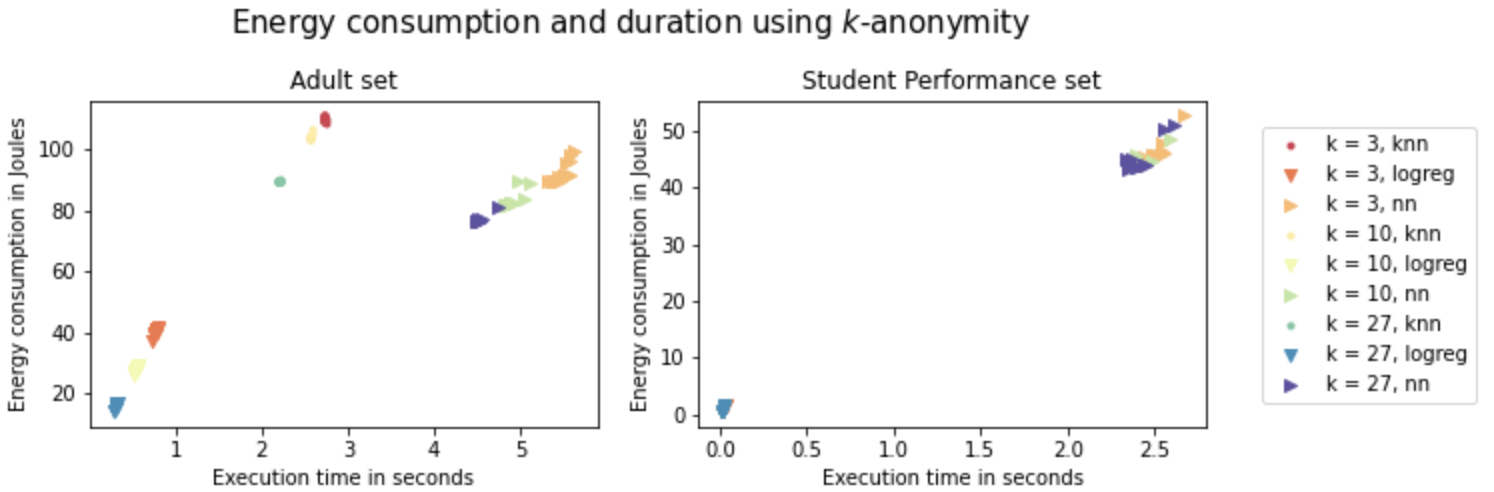}
    \caption{Scatter plot of the energy consumption in Joules and duration in seconds for the models trained on $k$-anonymity. The left plot shows the results of the models trained on the Adult set, the right plot shows the results of models trained on the Student Performance set.}
    \label{Paper:kanon}
\end{figure*}

\begin{figure*}[h]
    \centering
    \includegraphics[width=0.85\textwidth]{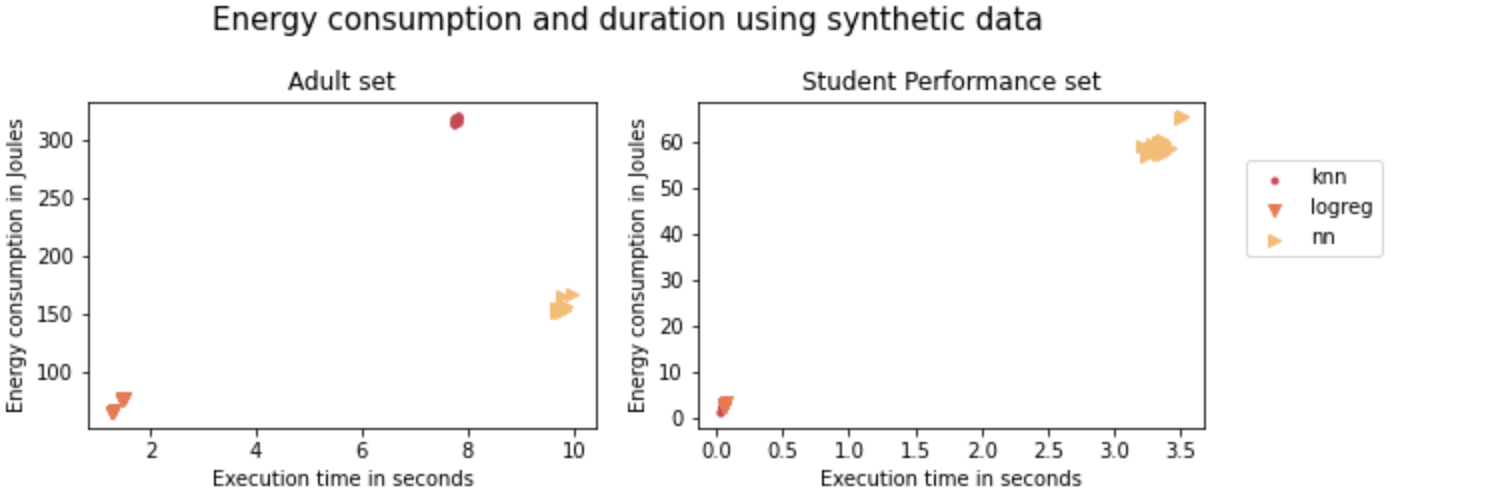}
    \caption{Scatter plot of the energy consumption in Joules and duration in seconds for the models trained on synthetic data. The left plot shows the results of the models trained on the Adult set, the right plot shows the results of models trained on the Student Performance set.}
    \label{Paper:SynthData}
\end{figure*}

In Table~\ref{PaperEnergies} we present the results of the energy consumption for the Adult data set on the left and the Student Performance set on the right. The benchmark is presented on the first row, after which the results are shown for the models trained on either anonymised or synthetic data. We present the deviation in percentages to the benchmark for these methods. The machine learning models are abbreviated to increase readability. In Figures~\ref{Paper:kanon} and~\ref{Paper:SynthData} we present a scatterplot of the data points for $k$-anonymity and synthetic data respectively. Complementary we provide the results of the Mann Whitney U test in Table~\ref{Paper:MannWhit}. Lastly, the idle energy consumption of the machine is measured. For each second of inactivity 7.512 Joules is consumed by the machine.

\begin{table*}[h!]
\centering
\caption{The average run time and energy consumption for the Adult and Student Performance data sets. On the rows we define the input data and on the columns we define each method, devised into run time and energy consumption. The benchmark resembles the unaltered data as obtained online, the percentages below show the deviation for each method to this benchmark.}
\label{PaperEnergies}
\begin{tabular}{lcccccclcccccc}
\toprule
\multicolumn{7}{c}{\textbf{Adult data set}} & \multicolumn{7}{c}{\textbf{Student Performance set}} \\
\cmidrule(r){1-7} \cmidrule(r){8-14}

    & \multicolumn{2}{c}{knn} & \multicolumn{2}{c}{LogReg} & \multicolumn{2}{c}{NN} &   & \multicolumn{2}{c}{knn} & \multicolumn{2}{c}{LogReg} & \multicolumn{2}{c}{NN} \\
\cmidrule(l{3pt}r{3pt}){2-3} \cmidrule(l{3pt}r{5pt}){4-5} \cmidrule(l{5pt}r{2pt}){6-7} \cmidrule(l{3pt}r{5pt}){9-10} \cmidrule(l{5pt}r{2pt}){11-12} \cmidrule(l{5pt}r{2pt}){13-14}
Benchmark & 8.19s & 329.33J & 1.41s & 73.70J & 9.85s & 157.59J & Benchmark & 0.04s & 2.16J & 0.06s & 2.95J & 3.40s & 61.20J \\
$k$=3 & -67\% & -67\% & -45\% & -46\% & -44\% & -42\% & $k$=3 & -25\% & -37\% & -66\% & -77\% & -26\% & -24\% \\
$k$=10 & -69\% & -68\% & -61\% & -62\% & -50\% & -47\% & $k$=10 & -75\% & -79\% & -66\% & -78\% & -28\% & -27\% \\
$k$=27 & -73\% & -73\% & -78\% & -78\% & -54\% & -51\% & $k$=27 & -75\% & -80\% & -66\% & -79\% & -29\% & -26\% \\
Synthetic data & -5\% & -4\% & -9\% & -10\% & -1\% & 0\% & Synthetic data & 0\% & +2\% & -33\% & +3\% & -1\% & -2\% \\
\bottomrule
\end{tabular}
\end{table*}

\begin{table*}[h!]
\centering
\caption{The results of the Mann Whitney U test applied to the energy consumption data. The presented p-value shows the probability of the method on the row being greater than the alternative in the column.}
\label{Paper:MannWhit}
\begin{tabular}{lcccclcccc}
\toprule
\multicolumn{5}{c}{\textbf{Adult data set}} & \multicolumn{4}{c}{\textbf{Student Performance set}} \\
\cmidrule(r){1-5} \cmidrule(r){6-10}
    & $k$=3 & $k$=10 & $k$=27 & Synthetic data &    & $k$=3 & $k$=10 & $k$=27 & Synthetic data \\
$k$=3 & x & p \textless 0.01 & p \textless 0.01 & p \textgreater 0.99 & $k$=3 & x & p \textless 0.01 & p \textless 0.01 & p \textgreater 0.99 \\
$k$=10 & p \textgreater 0.99 & x & p \textless 0.01 & p \textgreater 0.99 & $k$=10 & p \textgreater 0.99 & x & p \textless 0.01 & p \textgreater 0.99 \\
$k$=27 & p \textgreater 0.99 & p \textgreater 0.99 & x & p \textgreater 0.99 & $k$=27 & p \textgreater 0.99 & p \textgreater 0.99 & x & p \textgreater 0.99 \\
Synthetic & p \textless 0.01 & p \textless 0.01 & p \textless 0.01 & x & Synthetic & p \textless 0.01 & p \textless 0.01 & p \textless 0.01 & x \\
\bottomrule
\end{tabular}

\end{table*}

\subsection{Accuracy of the models}
Table~\ref{PaperAccModels} shows the accuracy score of the machine learning techniques for each data set and the privacy-enhancing technique. The benchmark reflects the accuracy scores for the unmodified or original data. To reflect the accuracies as accurate as possible we present the accuracy scores instead of the deviation to the benchmark. 

\begin{table}[]
\caption{The weighted accuracy over 10 measurements for each data set, machine learning technique and privacy-enhancing technique.}
\label{PaperAccModels}
\begin{tabular}{lcccccc}
\toprule
    & \multicolumn{3}{c}{\textbf{Adult data set}} & \multicolumn{3}{c}{\textbf{Student Performance set}} \\
\cmidrule(r){2-4} \cmidrule(r){5-7}
    & knn & LogReg & NN & knn & LogReg & NN \\
Benchmark & 0.820 & 0.846 & 0.846 & 0.700 & 0.719 & 0.706 \\
$k$=3 & 0.828 & 0.848 & 0.847 & 0.872 & 0.851 & 0.861 \\
$k$=10 & 0.832 & 0.842 & 0.842 & 0.826 & 0.826 & 0.826 \\
$k$=27 & 0.828 & 0.837 & 0.837 & 0.831 & 0.853 & 0.853 \\
Synthetic data & 0.802 & 0.828 & 0.828 & 0.728 & 0.756 & 0.755 \\
\bottomrule
\end{tabular}

\end{table}

\section{Discussion}
\label{sec:discussion}
First we will validate the presented benchmarks by comparing the results to related work. Then we discuss the results of the anonymisation process itself by looking at the energy consumption and suppression rate. Next we analyse (\textbf{RQ1}): the accuracy of the machine learning models trained on anonymised or synthetic data, after which we discuss (\textbf{RQ2}): the energy consumption of our machine learning models. Finally, we compare the results of $k$-anonymity with the results of synthetic data to formulate some guidelines on when to use which method.

\subsection{Validation of the benchmark}
In this section we compare the accuracy and energy consumption of our presented benchmark in Table~\ref{PaperAccModels} to the accuracy presented in other work. Since the goal of this paper is to examine the effect of anonymised data or synthetic data compared to original data, small differences in regard to other literature are deemed acceptable.

\subsubsection{\textbf{Adult data set}}
Starting with the accuracies for the Adult set, the k-nearest neighbours and logistic regression models outperform related work~\cite{hittmeir2019utility} with 0.6\% and 1.4\% respectively. The neural network has an accuracy that is 1.0\% lower compared to related work~\cite{chakrabarty2018statistical}. Since a neural network can have many hyperparameters that influence the accuracy we deem this difference acceptable.

\subsubsection{\textbf{Student Performance set}} Literature suggests that for k-nearest neighbours and logistic regression accuracies of 71.0\% and 76.0\% may be expected. Table~\ref{PaperAccModels} shows that our models deviate 1.0\% and 4.1\% respectively. As mentioned earlier, the focus of this paper lies on the difference between the benchmark and anonymised or synthetic data and thus we deem this difference acceptable. For the neural network, no work was found that used all columns for this task. The original paper reports an accuracy of 83.4\% but that included a larger network and hyperparameters that were optimised using grid search~\cite{cortez2008using}.\\

Finally, for the energy consumption. In general logistic regression is computationally efficient~\cite{ray2019quick}, which should result in a lower energy consumption. For both sets Table~\ref{PaperEnergies} shows that logistic regression consumes the least amount of energy compared to the other machine learning methods. Contrary to this finding, k-nearest neighbours becomes computationally more complex as the size of the data grows~\cite{ray2019quick}. Table~\ref{PaperEnergies} shows that indeed for the Adult set the k-nearest neighbours algorithm used $\sim$152 times more energy while the Adult set is just $\sim$17 times larger than the Student Performance set, confirming this statement. The energy consumption of neural network depends on many parameters~\cite{brownlee2021exploring}, which makes it hard to compare the baseline to literature.

\subsection{Creating k-anonymous and synthetic data}
In Table~\ref{PaperAnonymisationEnergy} we observe that creating synthetic data requires a higher run time than anonymising to obtain $k$-anonymity. In accordance with these findings we observe that the energy consumption is nearly 4 times higher for synthetic data. This is expected as the DataSynthesizer requires more steps to come to the new data than the ARX algorithm. We also find that the DataSynthesizer required more time for the Student Performance set than for the Adult data set, even though the Adult set contains more data as a whole. This might be explained by the DataSynthesizer having to analyse all features to create their respective probability distribution. Since the Student Performance set contains more attributes, as seen in Table~\ref{PaperDataOverview}, we hypothesise that this leads to more required run time for the Student Performance set.\\

An important note is that the privacy-enhancement process, either via generalisation and suppression or via synthetic data, is a marginal cost. The process requires only one run and then the data is anonymised and may be published. Running algorithms on this data might be done repeatedly by different users, resulting in an initial energy cost which is shared amongst users.

\subsection{Accuracy of the models}
For the accuracy of the models we divide our discussion in two parts; one component for the Adult set and another component for the Student Performance set. We study Table~\ref{PaperAccModels} in this section.

\subsubsection{\textbf{Adult data set}}
First of all, the amount of suppressed data. We see that for the larger Adult set the percentage of suppressed data lies between 19\%-32\%. For the Student Performance set the amount of suppressed data lies between 74\% - 83\%. As expected the smaller set is more sensitive to suppression than the larger set. Though we did not define suppression as information loss, it is an interesting finding. 
Secondly, when we look at the accuracies and compare the synthetic data to the benchmark we see a minor deviation of about -1.8\% accuracy for all machine learning methods. For the anonymised data using ARX these findings are different. The k-nearest neighbours algorithm actually obtains a higher accuracy for anonymised data than for the benchmark, with a maximum increase of 1.2\% for a $k$ of 10. For other $k$-values the increase is 0.8\% on the k-nearest neighbours algorithm. For both logistic regression and the neural network the accuracy tends to be similar ($k$=3) or a little lower with a maximum decrease of 0.9\%. We find a close resemblance between the accuracy of logistic regression and that of the neural network. Other literature finds a similar pattern for logistic regression and binary neural network~\cite{kazemnejad2010comparison, schumacher1996neural}.\\

\subsubsection{\textbf{Student Performance set}}
In the case of the models trained on the Student Performance set, the models trained on synthetic data have higher accuracy compared to models trained on the benchmark. The increase in accuracy ranges between 2.8\% for k-nearest neighbours and 4.9\% for the neural network. Other work suggests that this could be explained by clustering of synthetic data~\cite{hittmeir2019utility}. However, this increase is rather low compared to the accuracy of the models trained on anonymised data using $k$-anonymity. With increases up to 17.2\% the models trained with $k$-anonymity outperform the benchmark and synthetic data, regardless of the machine learning method. We tend to see the accuracy decrease as the value of $k$ increases.\\

The k-anonymised data sets improved the accuracy compared to the benchmark, probably due to a preselection of the data but this should be examined in future work. Apart from extending this work to more data sets to see if these findings endure, an interesting idea for further research would be to actively use feature selection in the anonymisation process. One could think of suppressing the columns with the least importance to the prediction task and then continue suppressing more important features if the set $k$-value requires so. This could lead to increased privacy with the least possible information loss.

\subsection{Energy consumption of training the ML models}
To discuss the energy consumption of the models trained on data with $k$-anonymity or on synthetic data, we again split our discussion into two parts; one for each data set. After analysis of both data sets we will try to generalise our findings. An important remark is that the presented values should be seen as indications of energy consumption. Their individual value should be related to other values to obtain a respective score.

\subsubsection{\textbf{Adult data set}}
For the energy consumption of models trained on the Adult data set, the left-side of Table~\ref{PaperEnergies} shows us that the synthetic data lies in close range compared to the benchmark. This indicates that the DataSynthesizer has similar characteristics to the original data. For the models trained on anonymised data using $k$-anonymity we see a clear deviation from the benchmark. Depending on the chosen $k$-value, the run time and energy consumption decrease. Similar to other work, we find that the run time and energy consumption decrease as we increase our $k$-value~\cite{oprescu2022energyk}. Additionally, Figure~\ref{Paper:kanon} shows that the measurements for models trained using $k$-anonymity are closely clustered around their mean. This indicates that there is little variation amongst the data. We see a similar trend in Figure~\ref{Paper:SynthData} for the models trained on synthetic data. Affirmative with these findings are the results of the Mann Whitney U test, presented in Table~\ref{Paper:MannWhit}. The outcomes in this table show that the probability of the higher $k$-values resulting in a higher energy cost are less than 0.01. Hence we can reject this hypothesis and confirm that the higher $k$-values lead to a lower energy consumption. Finally we see that the probability for synthetic data consuming more energy compared to any $k$-value is $>$~0.99. So we may conclude that models trained on synthetic data consume significantly more energy than the same models trained on anonymised data using $k$-anonymity.\\

\subsubsection{\textbf{Student Performance set}}
When we look at the right-side of Table~\ref{PaperEnergies} and compare it to our findings on the Adult set, we find that the smaller Student Performance set also has a lower run time and related lower energy consumption. This finding holds for the benchmark as well as for the anonymised and synthetic data. Similar to the Adult data, we observe that the synthetic data shows close resemblance to the benchmark. This similarity extends to the anonymised data, where we see both the run time and energy consumption decrease as the chosen $k$-value increases. In addition, Figures~\ref{Paper:kanon} and~\ref{Paper:SynthData} clearly show that the data points are clustered with little variation around their mean. As with the Adult data set, we see in Table~\ref{Paper:MannWhit} that the probability of higher $k$-values resulting in a higher energy cost are less than 0.01. Again the probability of models trained on synthetic data resulting in a higher energy consumption are $>$~0.99. 

For the neural network we see an interesting finding where the run time decreases faster than the energy consumption. Conversely, the k-nearest neighbours and logistic regression do show a trend for the run time and energy consumption. We therefore find that the run time and energy consumption might be correlated, but there is no definite relation between those. This could also be due to the energy consumption resembling the initial energy cost of the neural network.\\

Combining these findings to obtain a general view, we find that there is a relation between the chosen $k$-value and the run time and energy consumption of machine learning techniques. More specifically, as the chosen $k$ increases we find a significant decrease in run time and energy consumption. This is likely due to an increased amount of suppressed attributes but that does not account for all decline. In addition to this we find that machine learning techniques trained on synthetic data show close resemblance to the benchmark with respect to their run time and energy consumption.

\subsection{Comparing $k$-anonymity to synthetic data}
When comparing the Adult set to the Student Performance set we see some surprising results. Where the ML techniques trained on synthetic data show a lower accuracy for the Adult data set, they have a higher accuracy for the Student Performance set. Our findings coincide with earlier work~\cite{hittmeir2019utility} which states that the accuracy of synthetic data could be influenced by how DataSynthesizer clusters the data set, yet no evidence is provided. Apart from that, we observe that models trained on anonymised data using $k$-anonymity outperform the models trained on synthetic data for both data sets in terms of accuracy. This deviation is rather small (+0.018) for the Adult data set, but is more distinct for the Student Performance set (up to +0.172).

Another finding, regardless of the data and privacy-enhancing techniques, is that the machine learning methods logistic regression and the neural network show a close resemblance for both sets. This could lie in the fact that the neural network was made a binary classifier, which will be discussed in section D of the related work.

\subsection{Limitations of this study}
The use of two data sets in this work can be seen as a limitation; using more data sets of different properties such as size and data types could lead to insight into whether $k$-anonymity generally outperforms synthetic data or just for some specific data sets. However, as this research is still in an early phase of exploration, we adopt an incremental approach.\\

Moreover, when cleaning and preprocessing the data unique values were left out. This removal of outliers can be seen as bias, though this only occurred for one data subject in the entire Adult data set. Besides, this practise is compliant with literature~\cite{escobar2021quality}.

\section{Related work}
\label{sec:rel-work}
To the best of our knowledge, no work is conducted on comparing synthetic data with anonymised data with respect to machine learning looking at both accuracy and energy consumption. Below we will discuss some related work that served as inspiration for this research.

\subsection{Energy consumption versus accuracy}
A recent study discusses the trade-off between energy consumption and machine learning performance for neural networks~\cite{brownlee2021exploring}. Other recent work investigates the influence of the chosen $k$ value on energy consumption and accuracy~\cite{oprescu2022energyk}. This paper tries to broaden the latter research by including synthetic data as another privacy-enhancing technique.

\subsection{Machine learning performance on anonymised data}
Though a lot of research is conducted on both the proof and the influence of $k$-anonymity to privacy,
little research has been done regarding the influence of $k$-anonymity on the performance of machine learning. Recent research shows that by increasing the $k$ the information loss increases as well~\cite{slijepvcevic2021k}. Other studies strengthen this claim by showing that for k-nearest neighbours $k$-anonymity can outperform the baseline depending on the chosen value of $k$~\cite{oprescu2022energyk}. Hence a data set that holds $k$-anonymity is not always outperformed by its original parent data set.

\subsection{Machine learning performance on synthetic data}
As is the case with $k$-anonymity, research shows that the use of synthetic data can lead to information loss~\cite{ping2017datasynthesizer, hittmeir2019utility}. This does not mean that models trained on synthetic data will always yield a lower accuracy than models trained on the original data. For some data sets models trained on the synthetic data actually outperformed the models trained on original data~\cite{hittmeir2019utility}. Both works emphasise the need for more research in this field as the impact remains unclear.

The findings discussed in section C of the discussion show a similar trend. Compliant to other literature~\cite{hittmeir2019utility}, the models trained on synthetic data show a minor decrease in accuracy compared to the benchmark models trained using the original data.

\section{Conclusion}
\label{sec:conclusion}
The goal of this research was to examine the differences between machine learning trained on anonymised data using $k$-anonymity or on synthetic data, with a focus on information loss and energy consumption. To aim the research we defined two research questions:
\begin{enumerate}
    \item[\textbf{RQ1:}] Which method has the least information loss?
    \item[\textbf{RQ2:}] Which method is the least energy consuming?
\end{enumerate}
In the research method and experimental set-up we provide the approach to answer these questions.

We find that $k$-anonymisation via generalisation and suppression uses approximately a quarter of the energy compared to creating synthetic data. The run time of anonymisation is about half of creating synthetic data. However we also find that, especially for the smaller Student Performance set, quite some data is suppressed when using $k$-anonymity. For synthetic data no data is suppressed. We find that for k-nearest neighbours, logistic regression and a simple neural network the loss of accuracy is the least when the models are trained on anonymised data using $k$-anonymity. There is an indication that models trained on anonymised data can outperform models trained on the original data. 

Relating back to \textbf{RQ1}, we may conclude that models trained on anonymous data via $k$-anonymity have lower loss of accuracy than models trained on synthetic data. However we should be aware that generalisation and suppression leads to quite some suppression of the data set.

Then for the energy consumption, \textbf{RQ2}, the models trained on anonymous data have a lower run time and overall energy consumption than the models trained on the original or synthetic data. This is seen in both data sets across all the machine learning models. We think that most of the decrease in energy consumption can be related to the suppression of data. All in all we may conclude that regarding energy consumption we prefer $k$-anonymity over synthetic data.

Apart from the answers on the research questions above, we also find that anonymising data can lead to both a higher accuracy as well as a lower energy consumption. This finding is particularly interesting because these two important properties do not have to be a trade-off and exclude each other.

Ultimately, when going back to our research question, using $k$-anonymity over synthetic data could be preferable as it has a higher accuracy over both data sets and consumes less energy. However, this is also dependent on the size of the data set and on whether suppressing the data set is a decision factor.

\section*{Acknowledgement}
This work is partially funded through the AMdEX Fieldlab project supported by Kansen Voor West EFRO (KVW00309) and the province of Noord-Holland.

In addition to that, this work was financially supported by the Dutch NWO Research project “Data Logistics for Logistics Data” (DL4LD, www.dl4ld.net), supported by the Dutch Top consortia for Knowledge and Innovation “Institute for Advanced Logistics” (TKI Dinalog, www.dinalog.nl) of the Ministry of Economy and Environment in The Netherlands and the Dutch Commit-toData initiative (https://commit2data.nl/).

\newpage

\bibliography{literature}
\bibliographystyle{IEEEtran}

\end{document}